\documentclass{article}

\usepackage{arxiv}
\usepackage[utf8]{inputenc} 
\usepackage[T1]{fontenc}    
\usepackage{hyperref}       
\usepackage{url}            
\usepackage{booktabs}       
\usepackage{amsfonts}       
\usepackage{nicefrac}       
\usepackage{microtype}      
\usepackage{lipsum}
\usepackage{graphicx}
\graphicspath{ {./images/} }
\usepackage{amsmath}
\usepackage{multirow}
\usepackage{subcaption}

\title{Mam-App: A Novel Parameter-Efficient Mamba Model for Apple Leaf Disease Classification}
\author{%
\textbf{Md Nadim Mahamood}$^{1}$,
\textbf{Md Imran Hasan}$^{1}$,
\textbf{Md Rasheduzzaman}$^{1}$,
\textbf{Ausrukona Ray}$^{2}$\\
\textbf{Md Shafi Ud Doula}$^{3}$,
\textbf{Kamrul Hasan}$^{4}$\\[2mm]
$^{1}$Computer Science and Engineering, Begum Rokeya University, Rangpur, Bangladesh\\
$^{2}$Computer Science, Kumamoto University, Kumamoto, Japan\\
$^{3}$Information and Communication Technologies, Asian Institute of Technology, Pathum Thani, Thailand\\
$^{4}$Computer Science, Texas State University, San Marcos, Texas, USA\\[2mm]
\texttt{\{nadim649649, imranhasanmhs13, sobuj.bru, shafi.cse.brur\}@gmail.com}\\
\texttt{ray@st.cs.kumamoto-u.ac.jp},
\texttt{kamrul.hasan@txstate.edu}%
}
\renewcommand{\today}{}

\begin{document}
\maketitle

\begin{abstract}
The rapid growth of the global population, alongside exponential technological advancement, has intensified the demand for food production. Meeting this demand depends not only on increasing agricultural yield but also on minimizing food loss caused by crop diseases. Moreover, access to nutritious food is essential for maintaining a healthy life. Crop diseases significantly affect food security and the financial and economic stability of countries. Diseases account for a substantial portion of apple production losses, despite apples being among the most widely produced and nutritionally valuable fruits worldwide. Previous studies have employed machine learning techniques for feature extraction and early diagnosis of apple leaf diseases, and more recently, deep learning–based models have shown remarkable performance in disease recognition. However, most state-of-the-art deep learning models are highly parameter-intensive, resulting in increased training and inference time. Although lightweight models are more suitable for user-friendly and resource-constrained applications, they often suffer from performance degradation. To address the trade-off between efficiency and performance, we propose Mam-App, a parameter-efficient Mamba-based model for feature extraction and leaf disease classification. The proposed approach achieves competitive state-of-the-art performance on the PlantVillage Apple Leaf Disease dataset, attaining 99.58\% accuracy, 99.30\% precision, 99.14\% recall, and 99.22\% F1-score, while using only 0.051M parameters. This extremely low parameter count makes the model suitable for deployment on drones, mobile devices, and other low-resource platforms. To demonstrate the robustness and generalizability of the proposed model, we further evaluate it on the PlantVillage Corn Leaf Disease and Potato Leaf Disease datasets. The model achieves 99.48\%, 99.20\%, 99.34\%, and 99.27\% accuracy, precision, recall, and F1-score on the corn dataset, and 98.46\%, 98.91\%, 95.39\%, and 97.01\% on the potato dataset, respectively. Additionally, features extracted from the penultimate layer of the proposed model are visualized using Principal Component Analysis and t-distributed Stochastic Neighbor Embedding, showing clear class-wise clustering. We further validate the quality of the extracted features by training Random Forest and XGBoost classifiers, demonstrating the strong discriminative capability of Mamba-based representations.
\end{abstract}


\section{Introduction}

The continuous growth of the global population has heightened the demand for food, and the Office of the Director of National Intelligence projects that the world population will reach 9.2 billion by 2040\footnote{\url{https://www.dni.gov/files/ODNI/documents/assessments/GlobalTrends_2040.pdf}}. 
Furthermore, the Food and Agriculture Organization of the United Nations (FAO) estimates that global food production must increase by approximately 70\% by 2050 to meet future needs \footnote{\url{https://www.fao.org/newsroom/detail/2050-A-third-more-mouths-to-feed/}}. However, recent evidence shows that 2.3–2.9 billion people are unable to afford a healthy diet because their incomes are insufficient relative to the rising cost of nutritious foods \cite{stehl2025global}, highlighting a widening gap between food availability and economic access. To satisfy rising food demand, agricultural technologies are advancing rapidly; nevertheless, diseases and pests remain major obstacles. For example, the FAO reports that pests cause 20–40\% of global crop losses annually, imposing an estimated economic burden of about USD 290 billion.\footnote{\url{https://www.fao.org/newsroom/detail/New-standards-to-curb-the-global-spread-of-plant-pests-and-diseases/en}}

Apples are recognized as nutrient-dense fruits containing high levels of dietary fiber, antioxidants, vitamins, and minerals, which have been shown to play a role in the prevention of various chronic diseases \cite{boyer2004apple, li2024design} and one of the most productive fruits due to their high nutritional values \cite{khan2022deep}.  It is the third most produced fruit in the world \cite{de2022sustainable, lv2025comprehensive} and FAO reports that, in 2022, the global output of apples exceeded 95 million (M) metric tons \footnote{\url{https://www.fao.org/home/en/}} and due to  economic and social development and rise in health consciousness, it's now increasing  for selecting and breeding new, particular varieties \cite{chen2022deep}. However, apple trees are highly vulnerable to insects and pathogenic microorganisms \cite{ali2025fine}, making diseases a frequent and serious problem in apple production that leads to substantial economic losses \cite{fan2025semi} and spreads rapidly, often leading to significant decreases in crop yield and quality \cite{thomas2018benefits}.

Early diagnosis and accurate identification of apple leaf diseases are essential for controlling the spread of infection, minimizing yield loss, and ensuring the sustainable and healthy development of the apple industry \cite{jiang2019real, chao2020identification}. However, disease monitoring in orchards still relies heavily on manual inspection by experts, which is time-consuming, subjective, and prone to inconsistency and error \cite{madufor2017detection}. Manual apple sorting and disease inspection face additional challenges, including fruit collisions, surface damage, low grading efficiency, strong human influence, and high labor consumption \cite{yu2024field}. Monitoring large cultivation areas is tedious for farmers and requires substantial training, experience in symptom recognition, and broad knowledge of multiple plant diseases \cite{blancard2012tomato}. Moreover, variations in farmers’ skills and backgrounds lead to poor scientific consistency and limited reliability in manual inspection \cite{trivedi2021early}.

Traditional laboratory-based diagnostic techniques, such as Western blotting, enzyme-linked immunosorbent assay, and microarrays, can accurately identify plant pathogens \cite{yulita2023mobile}. However, their high cost, operational complexity, and delayed feedback make them impractical for real-time and large-scale agricultural deployment. On the other hand, imaging-based approaches, including visible, spectral, and fluorescence imaging, provide non-destructive and high-performance analysis. Nevertheless, these systems often rely on expert interpretation and require sophisticated, bulky, and expensive equipment, which limits scalability and field usability \cite{fan2025semi, jiang2019real, mahlein2013development, al2022artificial, hasan2022disease}. These limitations collectively highlight the need for efficient, low-cost, and automated apple disease detection systems suitable for real-world agricultural environments. Consequently, automatic apple disease classification technologies have attracted increasing attention and have become a key component of modern apple production, management, and market distribution systems \cite{ji2023apple}.

 Traditional machine learning (ML) and modern deep learning (DL) techniques have been widely applied to tomato plant disease analysis. In earlier studies, traditional ML approaches primarily relied on image processing and handcrafted feature–based analysis for disease recognition \cite{choudhary2020feature}. For instance, Singh et al \cite{singh2022extraction} proposed an apple leaf disease detection framework consisting of preprocessing, segmentation, feature extraction, and classification. In this approach, Gaussian filtering and Brightness Preserving Dynamic Fuzzy Histogram Equalization \cite{sheet2010brightness} were employed for noise reduction and contrast enhancement. The diseased regions were segmented using a region-based ROI extraction method that combined background removal and K-means clustering \cite{sinaga2020unsupervised}. Subsequently, handcrafted first- and second-order texture features were extracted and classified using different ML methods. Similarly, Chakraborty et al. \cite{chakraborty2021prediction} utilized image processing techniques such as histogram equalization \cite{patel2013comparative}, Otsu’s thresholding, and region-based image segmentation for feature extraction, which were then fed into a Support Vector Machine (SVM) \cite{suthaharan2016support} for apple disease classification. However, such image processing–based pipelines require manual feature design and parameter tuning, making them sensitive to illumination and background variations and limiting their robustness, scalability, and generalization when compared to end-to-end DL-based frameworks.

Deep learning has achieved significant success in both computer vision and natural language processing (NLP) applications, including agriculture (e.g., rice disease detection \cite{uddin2024e2etca}), due to its ability to model complex patterns in challenging environments. Similarly, deep learning is increasingly applied in apple to leaf disease analysis. For example,  Sujatha et al. \cite{sujatha2025advancing} propose an AI-based plant disease detection framework that combines CNN-based deep feature extraction (VGG19 \cite{simonyan2014very}, Inception-v3 \cite{szegedy2016rethinking}) with ML classifiers for Custard Apple Leaf. Again \cite{ccetiner2025applecnn} proposed a CNN-based apple leaf disease detection framework using multiple pre-trained architectures (VGG19, DenseNet169 \cite{huang2017densely}, MobileNetV2 \cite{sandler2018mobilenetv2}, Xception \cite{chollet2017xception}, and NASNetLarge \cite{zoph2018learning}) to enable early and accurate disease identification of apple leaf disease.  Recently, concepts from NLP, such as transformer attention \cite{vaswani2017attention}, have been applied in computer vision tasks as Vision Transformer (ViT) \cite{dosovitskiy2020image}, including apple leaf disease classification. For example, Huang et al. \cite{huang2025econv} proposed EConv-ViT, a robust apple leaf disease classification model that integrates ConvNeXt \cite{liu2022convnet} and ViT, enhanced with Efficient Channel Attention (ECA) and DropKey to capture both local and global features. On the other hand, Aboelenin et al. \cite{aboelenin2025hybrid} propose a hybrid CNN-ViT framework for plant leaf disease detection, including apple, where an ensemble of CNNs (VGG16, Inception-v3, DenseNet120) extracts global features, and the ViT captures local features. However, these deep learning models typically involve millions of parameters, requiring substantial training time and computational resources, which makes deployment on low-end devices challenging. Reducing the number of parameters can also degrade performance, making it difficult to maintain high accuracy with lightweight models.

Recently, Mamba \cite{gu2024mamba} has emerged as an alternative to Transformers, reducing the quadratic time complexity of self-attention to linear time while maintaining comparable performance. Its applications span NLP \cite{waleffe2024empirical}, computer vision \cite{rahman2024mamba}, medical imaging\cite{yue2024medmamba}, and agriculture\cite{ishwarya2025vision}, and in this study, we leverage this capability to accurately classify apple leaf diseases for early diagnosis. This optimized model also enables deployment on low-resource devices such as mobile phones, drone and robots. To the best of our knowledge, this is \textbf{the first work to apply Mamba for apple leaf disease classification}, and our contributions are as follows:

\begin{itemize}
    \item Proposed Mam-App, a parameter-efficient Mamba-based model for apple leaf disease classification, achieving 99.58\% accuracy, 99.30\% precision, 99.14\% recall, and 99.22\% F1-scor with only 0.051M parameters, suitable for low-resource devices.
    \item Demonstrated robustness and generalizability on PlantVillage Corn and Potato Leaf Disease datasets, achieving high accuracy, precision, recall, and F1-score.
    \item Visualized extracted features using Principle Component Analysis (PCA) and t-distributed Stochastic Neighbor Embedding (t-SNE), showing clear class-wise clustering, and validated their discriminative power with Random Forest and XGBoost classifiers.
\end{itemize}
\section{Related Work}
\subsection{Machine learning-based} 

Machine learning–based approaches for apple leaf disease detection often rely on handcrafted or color-based feature extraction methods, which are then fed into traditional classifiers, making them dependent on manually designed feature descriptors. Alqethami et al. \cite{alqethami2022disease} detect and classify apple leaf diseases by combining image processing, ML (SVM and KNN \cite{peterson2009k} with LBP features), and deep learning (modified GoogleNet \cite{szegedy2015going}). Their approach preprocesses and segments leaf images to remove background noise, extracts relevant features, and demonstrates that CNN-based classification achieves 98.5\% accuracy using GoogleNet, while SVM achieves 82.25\% on a dataset of 240 apple leaf images with four classes. Singh et al. \cite{singh2022extraction} propose a method for segmenting diseased regions in apple leaf images, particularly from varied and real-world backgrounds. Leaves are enhanced using Histogram Equalization, followed by a novel extraction algorithm to isolate diseased areas. From these segmented regions, color and texture features are extracted and classified as Marsonina coronaria or Apple scab using ML classifiers, achieving a best accuracy of 96.4\% with KNN. Similarly, Jayakrishna et al. \cite{jayakrishna2025utilizing} employ image processing to extract key features such as color, texture, and shape from apple leaf images and use the XGBoost algorithm to classify apple scab disease. On a dataset of 100 high-resolution apple leaf images, XGBoost achieved a test accuracy of 95.0\%, demonstrating its effectiveness for early disease detection and improved crop management.

\subsection{Deep learning-based}
Deep learning approaches are increasingly used in apple leaf disease detection for automatically learning hierarchical image features, with CNNs, attention-based models, and hybrid approaches outperforming traditional ML in accuracy and robustness. 
CNN \cite{o2015introduction} focuses on extracting spatial features from images using various filters to identify shapes, textures, and other patterns for classification. It is prominently used in tomato leaf disease detection. For example, Liu et al. \cite{liu2017identification} propose a deep learning–based approach for early and accurate identification of four common apple leaf diseases: Mosaic, Rust, Brown Spot, and Alternaria Leaf Spot, generating additional pathological images and designing a novel deep CNN architecture based on AlexNet \cite{krizhevsky2012imagenet}. Their model achieves 97.62\% accuracy, reduces model parameters by over 51.00 M, and demonstrates faster convergence and enhanced robustness on 13,689 diseased apple leaf images with a hold-out test set. Ali et al. \cite{ali2025fine} propose a fine-tuned EfficientNet-B0 \cite{tan2019efficientnet} CNN for automated classification of apple leaf diseases, incorporating architectural modifications including a global max pooling layer, dropout, regularization, and full-model fine-tuning, alongside a holistic training strategy with data augmentation, stratified splitting, class weighting, and transfer learning. Evaluated on the PlantVillage and Apple PV datasets, the model achieves 99.78\% and 99.69\% test accuracies, outperforming EfficientNet-B0, EfficientNet-B3, Inception-v3, ResNet50, and VGG16, while maintaining low memory consumption and FLOPs and demonstrating 11.00\% and 49.50\% accuracy improvements over EfficientNet-B0 on the APV and PV datasets, making it effective for resource-constrained environments. Additionally, Rohith et al. \cite{rohith2025integrated} employ CNN-based deep learning on a curated dataset of 3,200 apple leaf images across four classes (healthy, apple scab, black rot, cedar rust) and find that a Residual Neural Network achieves the highest validation accuracy of 98.9\%, outperforming VGG19, InceptionV3, EfficientNet, VGG16, MobileNet, and a baseline CNN, with an 80:20 training-to-validation split shown to yield optimal performance, highlighting the potential for real-time orchard management and early disease intervention.

Vision transformer \cite{dosovitskiy2020image} and hybrid models that combine CNN and attention mechanisms are increasingly popular for apple leaf disease detection. For example, Ullah et al. \cite{ullah2024efficient} develop AppViT, a hybrid vision model combining convolutional blocks with multi-head self-attention. The convolution blocks progressively encode local features, while stacked ViT blocks capture non-local dependencies and spatial patterns. Evaluated on the Plant Pathology 2021-FGVC8 dataset of 16,093 images (five classes: scab, healthy, frog-eye leaf spot, complex, rust), AppViT achieves 96.38\% precision, outperforming ResNet-50 and EfficientNet-B3 by 11.3\% and 4.3\%, respectively. To address class imbalance and enhance feature learning, extensive data augmentation (rotation, flipping, translation, brightness, zoom, contrast) was applied. Similarly, Huang et al. \cite{huang2025econv} propose EConv-ViT, which combines ConvNeXt \cite{liu2022convnet} and ViT with Efficient Channel Attention and DropKey for robust apple leaf disease recognition. Evaluated on lab- and field-captured images of healthy leaves, Alternaria blotch, grey spot, rust, and mosaic, the model achieves 99.20\% accuracy on lab images and 79.30\% on field images, outperforming ViT, ConvNeXt, and ResNet-50, and effectively capturing both local and global features for automated disease monitoring. Furthermore, Ali et al. \cite{math2025apple} propose L1-MHViT, a L1-regularized Multi-Head ViT that combines convolutional and transformer features. The model achieves 99.82\% accuracy, 99.46\% precision, 99.54\% recall, and 99.50\% F1-score on the Apple leaf dataset, outperforming baseline methods. Finally, Aboelenin et al. \cite{aboelenin2025hybrid} propose a hybrid CNN–ViT framework for plant leaf disease classification, combining an ensemble of CNNs (VGG16, Inception-v3, DenseNet120) for robust global feature extraction with a ViT for precise local feature learning. Evaluated on the PlantVillage Apple leaf dataset with four classes, the model achieves a high accuracy of 99.24\%, outperforming recently published methods.

Machine learning–based classification suffers from a lack of generalizability and depends heavily on handcrafted features \cite{uddin2024e2etca}. In contrast, deep learning achieves highly prominent performance in CNNs and attention-based models, but these come with a huge number of parameters, leading to high costs, long training times, large GPU power requirements, increased energy consumption, and a higher carbon footprint. Reducing parameters, however, can hinder performance. To address this, we propose Mam-App, which utilizes convolution and mamaba instead of attention to capture fine-grained features, 99.58\% accuracy, 99.30\% precision, 99.14\% recall, and 99.22\% F1-score with only 0.051M parameters on PlantVillage Apple Leaf Disease dataset. Additionally, to demonstrate the robustness and generalizability of the proposed model, we further evaluate it on the PlantVillage Corn Leaf Disease and Potato Leaf Disease datasets. The model achieves 99.48\%, 99.20\%, 99.34\%, and 99.27\% accuracy, precision, recall, and F1-score on the corn dataset, and 98.46\%, 98.91\%, 95.39\%, and 97.01\% on the Potato dataset, respectively.

\section{Materials and methods}

\begin{figure*}[!htbp]
\centering
\includegraphics[width=\linewidth]{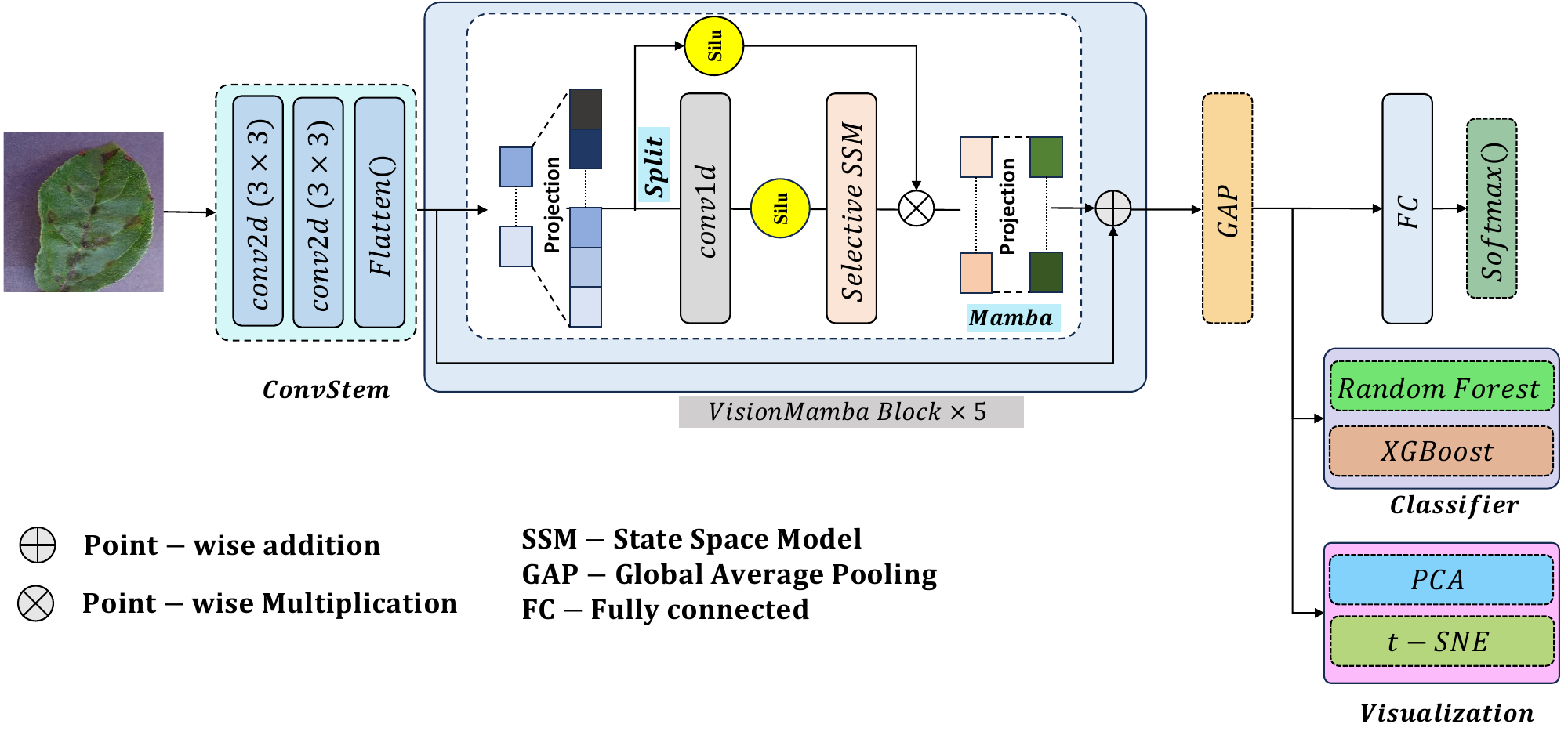}
\caption{Architecture of the proposed Mam-App model. The input image passes through two convolution layers for feature enrichment, followed by flattening and five VisionMamba blocks to capture local and global information. Global average pooling (GAP) aggregates features for classification via an FC layer with Softmax. The GAP features are also used for PCA/t-SNE visualization and Random Forest/XGBoost classification to demonstrate the strong feature extraction ability of Mam-App}
\label{fig:1_Mamba-app_architecture}
\end{figure*}

\subsection{Methods}

The proposed Mam-App model leverages the efficiency of the Mamba state-space model \cite{gu2024mamba} to achieve high-performance apple leaf disease classification with a lightweight architecture. The model is composed of several stages that progressively extract hierarchical features from the input image and transform them into discriminative representations. The overall architecture of the proposed approach is illustrated in Figure~\ref{fig:1_Mamba-app_architecture}.

\subsubsection{Stem Block}

The stem block serves as the initial feature extractor, responsible for capturing low-level spatial information from the input image. It consists of two consecutive $3 \times 3$ convolutional layers with strides $2$ and $2$, respectively, where each convolution is followed by batch normalization and GELU activation. This design effectively reduces the spatial resolution while preserving important visual cues such as edges, textures, and color variations.

Given an input image $X \in \mathbb{R}^{256 \times 256 \times 3}$, the stem block can be formulated as:

\begin{equation}
X_s = \text{Conv}{3 \times 3}^{(2)} \big( \text{BN}(\text{GELU}(\text{Conv}{3 \times 3}^{(2)}(X))) \big),
\end{equation}

where the first convolution produces feature maps of size $\mathbb{R}^{128 \times 128 \times 16}$, and the second convolution outputs $\mathbb{R}^{64 \times 64 \times 32}$. This stage efficiently reduces spatial dimensions while retaining essential visual information for subsequent processing.

\subsubsection{Feature Flattening}

The output of the stem block is reshaped into a sequence of tokens to enable sequence modeling via Mamba blocks. Specifically, the spatial dimensions are flattened and transposed:

\begin{equation}
X_t = \text{FlattenTranspose}(X_s) \in \mathbb{R}^{N \times C},
\end{equation}

where $N = H \times W$ denotes the number of tokens (e.g., $64 \times 64 = 4096$ for a $256 \times 256$ input image), and $C$ represents the channel dimension. For simplicity, the batch dimension is omitted in this formulation.

\subsubsection{VisionMamba Block}

The VisionMamba block \cite{zhu2024vision} constitutes the core computational unit of the Mam-App model. Each block consists of a layer normalization, a Mamba module, and a residual connection, enabling efficient modeling of both local and global dependencies across token sequences.

The output of the previous block, denoted as $X_t \in \mathbb{R}^{N \times C}$, is passed into the VisionMamba block. For an input size of $4096 \times 32$, the block computes:

\begin{equation}
X_{\text{out}} = X_{\text{in}} + \text{Mamba}(\text{LN}(X_t)),
\end{equation}

where $\text{LN}(\cdot)$ denotes layer normalization and the residual connection ensures stable gradient flow.

The \textbf{Mamba module} performs token-wise feature transformation and selective state-space modeling, described as:

\begin{equation}
\begin{aligned}
X_{\sim} &= \text{Linear}_{\text{in}}\big(\text{LN}(X_{\text{t}})\big) \in \mathbb{R}^{N \times 2 d_{\text{inner}}}, \\
X_{\text{token}}, Z &= \text{Split}(X_{\sim}, \text{dim}=2), \quad X_{\text{token}}, Z \in \mathbb{R}^{N \times d_{\text{inner}}}.
\end{aligned}
\end{equation}

The dimension of $X_{\sim}$, $X_{\text{token}}$, and $Z$ is $4096 \times 64$, $4096 \times 32$, and $4096 \times 32$, respectively.  

These tokens $X_{\text{token}}$ and $Z$ are then used in subsequent Mamba operations, including depthwise convolution, selective state-space modeling, gating, and final projection.

The token $X_{\text{token}}$ is further processed through depthwise convolution and selective state-space modeling:
\begin{equation}
\begin{aligned}
X_{\text{conv}} &= \text{SiLU}\Big(\text{Conv1D}_{\text{depthwise}}(X_{\text{token}})\Big), \\
Y_n &= \text{Selective SSM}(X_{\text{conv}}),
\end{aligned}
\end{equation}

where $X_{\text{conv}} \in \mathbb{R}^{N \times d_{\text{inner}}}$ is the depthwise-convolved ($ \text{kernel size} = 4, \text{stride} =1$) token representation after SiLU activation, and $Y_n \in \mathbb{R}^{N \times d_{\text{inner}}}$ is the output of the selective state-space model, capturing both local and global dependencies. In our case, $X_{\text{conv}}$ and $Y_n$ have dimensions $4096 \times 32$.

This output $Y_n$ is further combined with the second half of the split token, $Z$, through gating and a final linear projection:

\begin{equation}
\begin{aligned}
Y_{\text{gated}} &= Y_n \odot \text{SiLU}(Z), \quad Y_{\text{gated}} \in \mathbb{R}^{4096 \times 32}, \\
X_{\text{Mamba}} &= \text{Linear}_{\text{out}}(Y_{\text{gated}}), \quad X_{\text{Mamba}} \in \mathbb{R}^{4096 \times 32}.
\end{aligned}
\end{equation}

Finally, a residual connection adds the input $X_{\text{in}}$ to the Mamba output to form the final block output:

\begin{equation}
X_{\text{out}} = X_{\text{in}} + X_{\text{Mamba}} \quad \text{(residual connection)}.
\end{equation}

After five consecutive VisionMamba blocks, the output $X_{\text{out}} \in \mathbb{R}^{4096 \times 32}$ is passed to the next stage of the model for global pooling and classification.

\subsubsection{Classifier Head}

The classifier head aggregates the refined token features and produces the final disease predictions. First, layer normalization is applied to the output tokens of the last VisionMamba block:

\begin{equation}
X_{\text{LN}} = \text{LN}(X_{\text{out}}), \quad X_{\text{LN}} \in \mathbb{R}^{4096 \times 32}.
\end{equation}

Global average pooling (GAP) is then applied across the token dimension to summarize the sequence into a compact feature representation:

\begin{equation}
X_{\text{GAP}} = \text{GAP}(X_{\text{LN}}) \in \mathbb{R}^{32}.
\end{equation}

Finally, a fully connected (FC) layer maps the aggregated feature vector to the target class space:

\begin{equation}
X_{\text{class}} = \text{FC}(X_{\text{GAP}}) \in \mathbb{R}^{4},
\end{equation}

\begin{equation}
X_{\text{Pred}} = \text{Softmax}(X_{\text{class}})
\end{equation}

where $X_{\text{Pred}} \in \mathbb{R}^{4}$ represents the predicted class probabilities for the four apple leaf disease categories. The class associated with the maximum probability value is selected as the final predicted class.

To further evaluate the discriminative power of the learned representations, the pooled feature vector $X_{\text{GAP}} \in \mathbb{R}^{32}$ is additionally used as input to classical ML classifiers, including Random Forest and Extreme Gradient Boosting (XGBoost), as part of an ablation study. In this setting, the VisionMamba backbone serves as a feature extractor, while the final classification layer is replaced by the respective machine learning classifier. This analysis highlights the effectiveness and generalization capability of the learned feature embeddings independent of the neural classifier head.

\subsection{Materials}

\subsubsection{Dataset}

\begin{figure*}[!t]
\centering

\textbf{Apple leaf disease dataset (4 classes)}\\[4pt]

\begin{minipage}{0.23\linewidth}
    \centering
    \includegraphics[width=\linewidth]{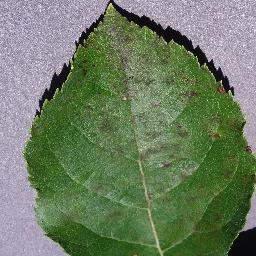}
    \subcaption{Apple Scab}
\end{minipage}
\hfill
\begin{minipage}{0.23\linewidth}
    \centering
    \includegraphics[width=\linewidth]{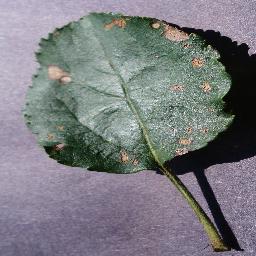}
    \subcaption{Apple Black Rot}
\end{minipage}
\hfill
\begin{minipage}{0.23\linewidth}
    \centering
    \includegraphics[width=\linewidth]{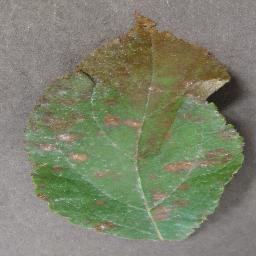}
    \subcaption{Apple Cedar Rust}
\end{minipage}
\hfill
\begin{minipage}{0.23\linewidth}
    \centering
    \includegraphics[width=\linewidth]{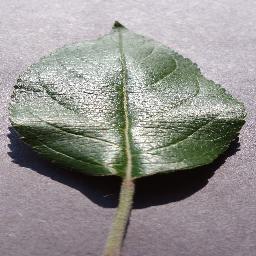}
    \subcaption{Apple Healthy}
\end{minipage}

\par\medskip

\textbf{Corn leaf disease dataset (4 classes)}\\[4pt]

\begin{minipage}{0.23\linewidth}
    \centering
    \includegraphics[width=\linewidth]{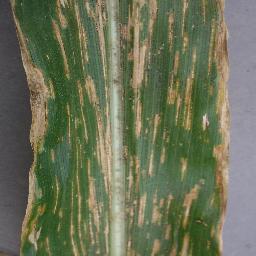}
    \subcaption{Corn Cercospora Leaf Spot}
\end{minipage}
\hfill
\begin{minipage}{0.23\linewidth}
    \centering
    \includegraphics[width=\linewidth]{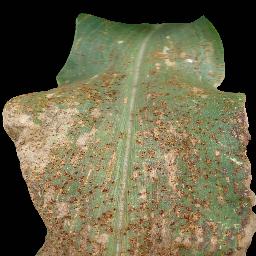}
    \subcaption{Corn Common Rust}
\end{minipage}
\hfill
\begin{minipage}{0.23\linewidth}
    \centering
    \includegraphics[width=\linewidth]{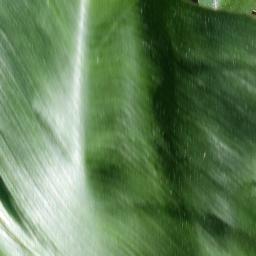}
    \subcaption{Corn Healthy}
\end{minipage}
\hfill
\begin{minipage}{0.23\linewidth}
    \centering
    \includegraphics[width=\linewidth]{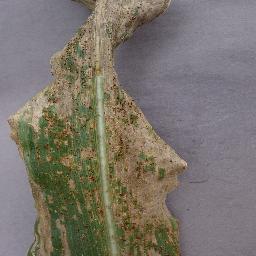}
    \subcaption{Corn Northern Leaf Blight}
\end{minipage}
\par\medskip

\textbf{Potato leaf disease dataset (3 classes)}\\[4pt]

\begin{minipage}{0.23\linewidth}
    \centering
    \includegraphics[width=\linewidth]{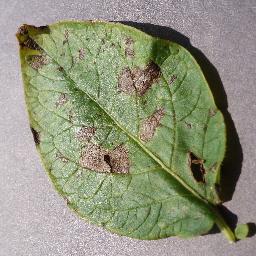}
    \subcaption{Potato Early Bligh}
\end{minipage}
\hfill
\begin{minipage}{0.23\linewidth}
    \centering
    \includegraphics[width=\linewidth]{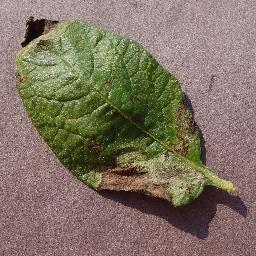}
    \subcaption{Potato Late Blight}
\end{minipage}
\hfill
\begin{minipage}{0.23\linewidth}
    \centering
    \includegraphics[width=\linewidth]{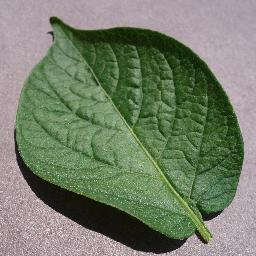}
    \subcaption{Potato Healthy}
\end{minipage}

\caption{Example image instances from the three crop leaf disease datasets used in this study. Each row corresponds to a different crop dataset, illustrating representative samples from all disease classes: Apple (4 classes), Corn (4 classes), and Potato (3 classes).}
\label{fig:dataset_examples}
\end{figure*}

To evaluate the effectiveness and generalization ability of the proposed Mam-App model, experiments were conducted on three publicly available plant disease datasets derived from the PlantVillage\footnote{https://www.kaggle.com/datasets/emmarex/plantdisease} repository, namely Apple Leaf Disease, Corn (Maize) Leaf Disease, and Potato Leaf Disease datasets. PlantVillage is a large-scale benchmark dataset containing high-quality RGB images of healthy and diseased plant leaves captured under controlled conditions.  The Apple Leaf Disease and Corn Leaf Disease datasets contain four classes, and the Potato Leaf Disease dataset contains three disease categories.

All images were resized to $256 \times 256$ pixels and normalized before being fed into the network. To enhance robustness and reduce overfitting, data augmentation was applied during training using random horizontal and vertical flipping, random rotation within $10^\circ$, and brightness jittering. The transformation pipeline is defined as:

\begin{equation}
\begin{aligned}
\texttt{Resize(256,256)} &\rightarrow \texttt{RandomHorizontalFlip} \rightarrow \texttt{RandomVerticalFlip} \\
&\rightarrow \texttt{RandomRotation(10)} \rightarrow \texttt{ColorJitter(0.3)} \rightarrow \texttt{ToTensor}.
\end{aligned}
\end{equation}

Each dataset was split into training, validation, and testing subsets while preserving the class distribution as much as possible. Approximately 70.00\% of the samples were used for training, 15.00\% for validation, and the remaining 15.00\% for testing. Details of each dataset are described below.

\textbf{\textit{Apple Leaf Disease Dataset:}}
The Apple dataset consists of four categories: \textit{Apple Scab}, \textit{Black Rot}, \textit{Cedar Apple Rust}, and \textit{Healthy}. A total of 3,171 images were used and divided into 2,218 for training, 474 for validation, and 479 for testing. The dataset exhibits class imbalance, where the healthy class dominates the distribution, while Cedar Apple Rust contains the fewest samples. The maximum-to-minimum class ratio is approximately 5.98, highlighting the challenging nature of the classification task. The distribution is shown in Table \ref{tab:apple_dist}.

\begin{table}[h]
\centering
\caption{Class distribution of the Apple Leaf Disease dataset.}
\label{tab:apple_dist}
\begin{tabular}{lcccc}
\hline
Class & Train & Val & Test & Total \\
\hline
Apple Scab & 441 & 94 & 95 & 630 \\
Black Rot & 434 & 93 & 94 & 621 \\
Cedar Apple Rust & 192 & 41 & 42 & 275 \\
Healthy & 1151 & 246 & 248 & 1645 \\
\hline
Total & 2218 & 474 & 479 & 3171 \\
\hline
\end{tabular}
\end{table}

\textbf{\textit{Corn (Maize) Leaf Disease Dataset:}}
The Corn dataset includes four categories: \textit{Cercospora Leaf Spot (Gray Leaf Spot)}, \textit{Common Rust}, \textit{Northern Leaf Blight}, and \textit{Healthy}. A total of 3,852 images were utilized, split into 2,695 training, 575 validation, and 582 testing samples. The most frequent class is Common Rust, while Cercospora Leaf Spot contains the fewest samples, producing a maximum-to-minimum class ratio of approximately 2.32. The distribution is shown in Table \ref{tab:corn_dist}.

\begin{table}[h]
\centering
\caption{Class distribution of the Corn (Maize) Leaf Disease dataset.}
\label{tab:corn_dist}
\begin{tabular}{lcccc}
\hline
Class & Train & Val & Test & Total \\
\hline
Cercospora Leaf Spot & 359 & 76 & 78 & 513 \\
Common Rust & 834 & 178 & 180 & 1192 \\
Northern Leaf Blight & 689 & 147 & 149 & 985 \\
Healthy & 813 & 174 & 175 & 1162 \\
\hline
Total & 2695 & 575 & 582 & 3852 \\
\hline
\end{tabular}
\end{table}

\textbf{\textit{Potato Leaf Disease Dataset:}}
The Potato dataset contains three classes: \textit{Early Blight}, \textit{Late Blight}, and \textit{Healthy}. In total, 2,152 images were used, with 1,506 for training, 322 for validation, and 324 for testing. The largest classes are Early Blight and Late Blight, while the healthy class has the fewest samples, resulting in a maximum-to-minimum class ratio of approximately 6.58. The distribution is shown in Table \ref{tab:potato_dist}.

\begin{table}[h]
\centering
\caption{Class distribution of the Potato Leaf Disease dataset.}
\label{tab:potato_dist}
\begin{tabular}{lcccc}
\hline
Class & Train & Val & Test & Total \\
\hline
Early Blight & 700 & 150 & 150 & 1000 \\
Late Blight & 700 & 150 & 150 & 1000 \\
Healthy & 106 & 22 & 24 & 152 \\
\hline
Total & 1506 & 322 & 324 & 2152 \\
\hline
\end{tabular}
\end{table}

\textbf{\textit{Discussion on Class Imbalance:}}
All three datasets present varying degrees of class imbalance, with the Apple dataset being the most skewed. The ratio between the largest and smallest classes is 5.98 for Apple, 2.32 for Corn, and 6.58 for Potato. Such imbalance can bias learning toward majority classes. The proposed Mam-App model mitigates this issue through robust feature extraction and augmentation strategies, enabling stable performance across minority and majority classes without requiring explicit resampling techniques.

\subsubsection{Training details}
The proposed Mam-App model was trained for $1000$ epochs with a batch size of $16-32$, using input images resized to $256 \times 256 \times 3$.
The model was trained using Cross-Entropy loss with label smoothing \cite{szegedy2016rethinking} set to 0.1 to reduce overconfidence in predictions. Label smoothing works by assigning a slightly lower probability to the correct class and distributing the remaining probability across the other classes. For instance, with a smoothing value of 0.1, the correct class receives a probability of 0.9, while the remaining 0.1 is distributed among the other classes. This technique helps improve generalization and reduces overfitting. Additional training details, including optimizer, weight decay, and hardware specifications, are summarized in Table \ref{tab:training_details}.

\begin{table}[!t]
\centering
\caption{Training configuration details used for model training.}
\renewcommand{\arraystretch}{1.3}
\begin{tabular}{l l}
\hline
\textbf{Parameter} & \textbf{Value} \\ \hline
Learning Rate (LR) & $1 \times 10^{-3}$ \\ \hline
Loss Function & Cross-entropy (label\_smoothing = 0.1) \\ \hline
Optimizer & AdamW (weight\_decay = $1 \times 10^{-5}$) \\ \hline
Batch Size & 16-32 \\ \hline
Input Image Size & (256, 256, 3) \\ \hline
Activation Function & Softmax \\ \hline
Hardware & NVIDIA GeForce RTX 3090 GPU\\ \hline
\end{tabular}
\label{tab:training_details}
\end{table}

\subsubsection{Performance evaluation metrics}

We consider four evaluation metrics, along with the number of parameters of the model, to validate the performance of the proposed approach. These metrics include Accuracy, Precision, Recall, and F1-score. Accuracy represents the ratio of correctly classified instances to the total number of predictions. Precision measures the proportion of correctly identified positive samples among all predicted positive samples. For example, when computing the precision of the healthy class, we treat healthy as the positive class and the remaining classes as negative; thus, precision indicates how many predicted healthy instances are actually healthy. Recall, in contrast, measures how many true positive samples are correctly identified among all actual positive samples. Following the same example, the recall of the healthy class reflects how many truly healthy instances are correctly classified as healthy. The F1-score is then computed as the harmonic mean of precision and recall, providing a balanced assessment of model performance. 

Since one of our datasets is imbalanced, we report the micro-averaged Precision, Recall, and F1-score to ensure equal contribution of each class to the final evaluation. Table~\ref{tab:evaluation_metric} presents the mathematical formulations of these metrics.

\begin{table} [!t]
\centering
\caption{Micro-averaged metrics used to evaluate the proposed Mam-App}
\label{tab:evaluation_metric}
\begin{tabular}{lc} \hline 
Metric & Expression  \\
\hline \\

Accuracy & 
$\displaystyle \frac{\sum_{i=1}^{K} \mathrm{TP}_i}{\sum_{i=1}^{K} (\mathrm{TP}_i + \mathrm{FP}_i + \mathrm{FN}_i)}$
\\ \\

Recall &
$\displaystyle \frac{\sum_{i=1}^{K}\mathrm{TP}_i}{\sum_{i=1}^{K}(\mathrm{TP}_i+\mathrm{FN}_i)}$
\\ \\

Precision &
$\displaystyle \frac{\sum_{i=1}^{K}\mathrm{TP}_i}{\sum_{i=1}^{K}(\mathrm{TP}_i+\mathrm{FP}_i)}$
\\ \\

F1-score  &
$\displaystyle
\frac{2 \times \Big(\sum_{i=1}^{K}\mathrm{TP}_i\Big)}
{2 \times \Big(\sum_{i=1}^{K}\mathrm{TP}_i\Big) 
+ \sum_{i=1}^{K}\mathrm{FP}_i + \sum_{i=1}^{K}\mathrm{FN}_i}$
\\ \\

\hline 
\end{tabular} 
\end{table}

\section{Experiment \& Results}

\subsection{Comparison with SOTA}

\begin{table*}[t]
\centering
\caption{Comparison with Existing Apple Leaf Disease Classification Methods. All measure in percentage and parameters in Million}
\label{tab:comparison}
\renewcommand{\arraystretch}{1.2}

\begin{tabular}{l l c c c c c}

\hline
\textbf{Ref.} & \textbf{Model} & \textbf{Params (M)} & \textbf{Accuracy (\%)} & \textbf{Precision (\%)} & \textbf{Recall (\%)} & \textbf{F1-score (\%)} \\

\hline
\multirow{2}{*}{\cite{vishnoi2022detection}}
& MobileNetV2        & 3.54  & 97.00   & 96.00   & 93.00   & 94.00 \\
& Deep CNN           & 0.12  & 98.00   & 98.00   & 97.00   & 97.00 \\

\hline
\cite{khan2019optimized}
& SVM                & N/A   & 97.20  & 97.00  & 97.15  & --   \\

\hline
\multirow{2}{*}{\cite{aboelenin2025hybrid}}
& VGG16 + Inception-v3 + & \multirow{2}{*}{268.0} 
& \multirow{2}{*}{99.24} 
& \multirow{2}{*}{99.00} 
& \multirow{2}{*}{99.00} 
& \multirow{2}{*}{99.00} \\
& DenseNet201 + ViT &  &  &  &  &  \\

\hline
\cite{rehman2021recognizing}
& RCNN + ResNet-50    &  25.6 & 96.60  & 96.75  & 96.00  & --   \\

\hline
\multirow{2}{*}{\cite{zeynalov2025automated} }
& EfficientNet-B4     & 19.0 & 93.08  & 90.74  & 91.04  & 90.62 \\
& MobileNet-V3-Large  &  5.4  & 93.71  & 94.73  & 93.17  & 93.91 \\

\hline
\multirow{2}{*}{\cite{kunduracioglu2024cnn}} 
& DenseNet121        &  8.0  & 99.79 & 99.74 & 99.90 & 99.82 \\
& Xception           &  22.9 & 99.79 & 99.42 & 99.74 & 99.57 \\

\hline
\textbf{Ours} & \textbf{Mam-App} & \textbf{0.051} & \textbf{99.58} & \textbf{99.30} & \textbf{99.14} & \textbf{99.22} \\

\hline
\end{tabular}
\end{table*}

\subsubsection{Performance Comparison with Existing Methods using PlantVillage apple dataset}

The proposed Mam-App model was evaluated against several state-of-the-art approaches for apple leaf disease classification, as summarized in Table~\ref{tab:comparison}. The comparison includes the number of parameters, Accuracy, Precision, Recall, and F1-score. Parameter counts for competing models were obtained from the original publications when available; otherwise, they were estimated using the official implementations.

As shown in Table~\ref{tab:comparison}, Mam-App achieves highly competitive performance while maintaining an extremely compact architecture with only 0.051M parameters. Despite its lightweight design, Mam-App attains an Accuracy of 99.58\%, Precision of 99.30\%, Recall of 99.14\%, and an F1-score of 99.22\%, outperforming several heavy CNN backbones such as DenseNet121 and Xception, which contain millions of parameters. For instance, Xception employs approximately 22.9M parameters, which is nearly 450$\times$ larger than Mam-App, yet offers only marginal improvements in classification metrics.

Compared with mobile-friendly architectures such as MobileNetV2 and MobileNet-V3-Large, Mam-App consistently provides higher Accuracy and F1-score while using orders of magnitude fewer parameters. Furthermore, although some recent hybrid and very deep models report near-perfect performance, these architectures sacrifice efficiency and deployability due to their large memory footprint and computational cost. In contrast, Mam-App is specifically designed as a parameter-efficient Mamba-based model, enabling robust feature representation through gated state-space mixing while remaining suitable for real-world and resource-constrained agricultural applications.

Overall, these results demonstrate that Mam-App effectively balances accuracy and efficiency, making it a practical and scalable solution for apple leaf disease recognition compared with existing state-of-the-art methods.

\subsection{Ablation study}

\subsubsection{Performance of Mam-App on the PlantVillage Apple, Corn, and Potato Datasets}
The proposed App-Mamba model performs direct end-to-end classification and demonstrates excellent performance on the PlantVillage Apple dataset, achieving an accuracy of 99.58\%, precision of 99.30\%, recall of 99.14\%, and an F1-score of 99.22\%, as reported in Table~\ref{tab:mamba_app_results_percent}. Notably, the feature-based variants, App-Mamba + Random Forest and App-Mamba + XGBoost, exhibit slightly lower performance, with accuracies of 99.58\% and 99.37\%, respectively. This indicates that the features learned by App-Mamba are highly discriminative and largely sufficient for accurate classification without requiring additional classifiers.

To evaluate the robustness and generalizability of the learned feature representations, we further test the model on the PlantVillage Corn and Potato datasets. Features extracted from the final layer of App-Mamba are provided as input to classical ML classifiers, namely Random Forest and XGBoost On the Corn dataset, App-Mamba achieves an accuracy of 99.48\%, with corresponding precision, recall, and F1-score values of 99.20\%, 99.34\%, and 99.27\%, respectively. The feature-based Random Forest and XGBoost classifiers yield slightly lower but comparable results, both attaining an accuracy of 99.31\%, demonstrating that the extracted features retain strong discriminative capability across different classification paradigms.

\begin{figure*}[!t]
\centering
\begin{minipage}{0.495\linewidth}
    \centering
    \includegraphics[width=\linewidth]{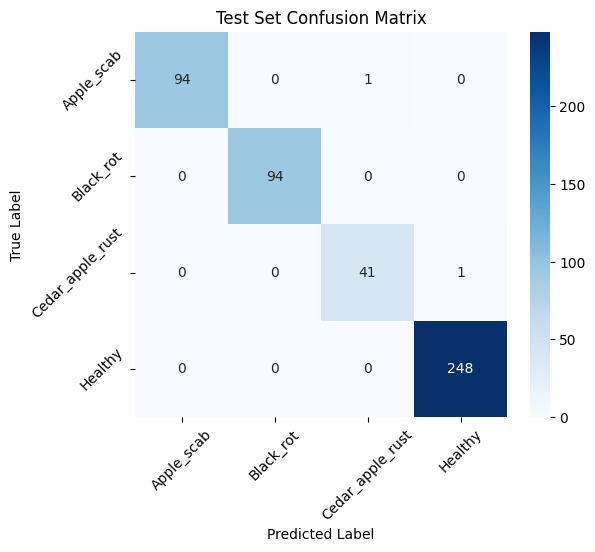}
    \subcaption{Apple}
    \label{fig:4a_Apple_confusion_matrix}
\end{minipage}
\hfill
\begin{minipage}{0.495\linewidth}
    \centering
    \includegraphics[width=\linewidth]{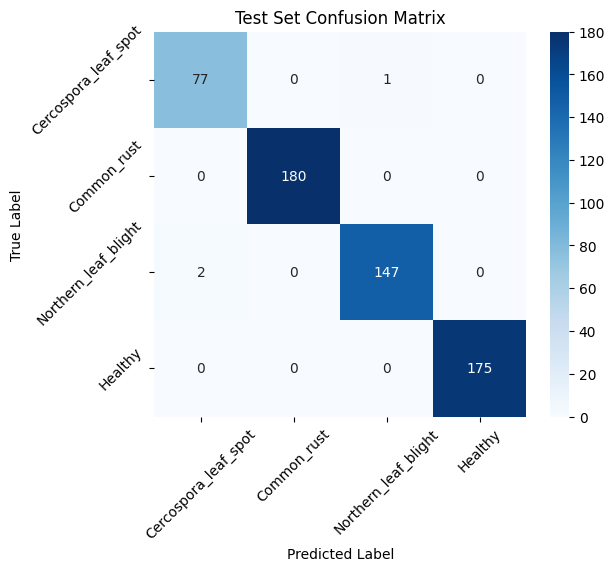}
    \subcaption{Corn}
    \label{fig:4b_Corn_confusion_matrix}
\end{minipage}
\hfill
\begin{minipage}{0.495\linewidth}
    \centering
    \includegraphics[width=\linewidth]{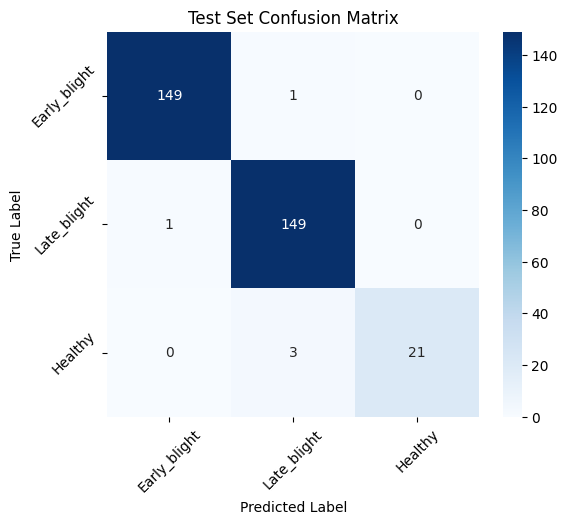}
    \subcaption{Potato}
    \label{fig:4c_Potato_confusion_matrix}
\end{minipage}

\caption{Confusion matrices for the Apple, Corn, and Potato datasets from PlantVillage, evaluated on the test set only.}
\label{fig:conf_matrix}
\end{figure*}

On the Potato dataset, App-Mamba attains an accuracy of 98.46\%, while the feature-based Random Forest and XGBoost classifiers achieve accuracies of 97.84\% and 98.15\%, respectively. These findings indicate that although the Potato dataset exhibits higher inter-class similarity, the learned representations remain informative and can be effectively transferred to downstream classifiers. Overall, the results confirm that App-Mamba learns robust and generalizable feature representations suitable for both direct classification and classical ML–based inference. The confusion matrices obtained using App-Mamba for the Apple, Corn, and Potato datasets are presented in Figure~\ref{fig:conf_matrix}.

\begin{table*}[t]
\centering
\caption{Performance comparison of the proposed Mamba-App model and its feature-based variants using Random Forest and XGBoost on Apple, Corn, and Potato datasets (percentage format).}

\label{tab:mamba_app_results_percent}
\begin{tabular}{llcccc}

\hline
\textbf{Dataset} & \textbf{Model} & \textbf{Accuracy (\%)} & \textbf{Precision (\%)} & \textbf{Recall (\%)} & \textbf{F1-score (\%)} \\

\hline
\multirow{3}{*}{Apple} 
& App-Mamba & 99.58 & 99.30 & 99.14 & 99.22 \\
& App-Mamba + Random Forest & 99.58 & 99.14 & 99.14 & 99.14 \\
& App-Mamba + XGBoost & 99.37 & 98.88 & 98.88 & 98.88 \\

\hline
\multirow{3}{*}{Corn} 
& App-Mamba & 99.48 & 99.20 & 99.34 & 99.27 \\
& App-Mamba + Random Forest & 99.31 & 98.89 & 99.18 & 99.03 \\
& App-Mamba + XGBoost & 99.31 & 98.89 & 99.18 & 99.03 \\

\hline
\multirow{3}{*}{Potato} 
& App-Mamba & 98.46 & 98.91 & 95.39 & 97.01 \\
& App-Mamba + Random Forest & 97.84 & 97.16 & 94.94 & 95.99 \\
& App-Mamba + XGBoost & 98.15 & 97.38 & 95.17 & 96.21 \\

\hline
\end{tabular}
\end{table*}

\subsubsection{Feature extract and visualization using the PlantVillage Apple, Corn, and Potato Datasets}

\begin{figure*}[!t]
\centering
\begin{minipage}{0.32\linewidth}
    \centering
    \includegraphics[width=\linewidth]{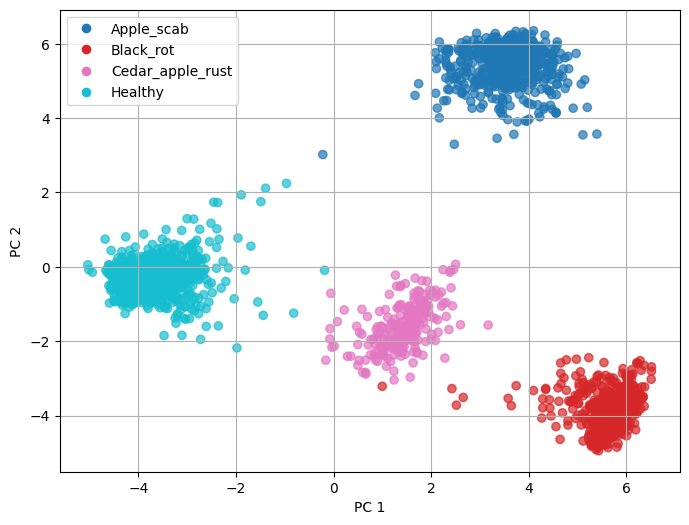}
    \subcaption{Apple 2D PCA}
    \label{fig:apple_2d_pca}
\end{minipage}
\hfill
\begin{minipage}{0.32\linewidth}
    \centering
    \includegraphics[width=\linewidth]{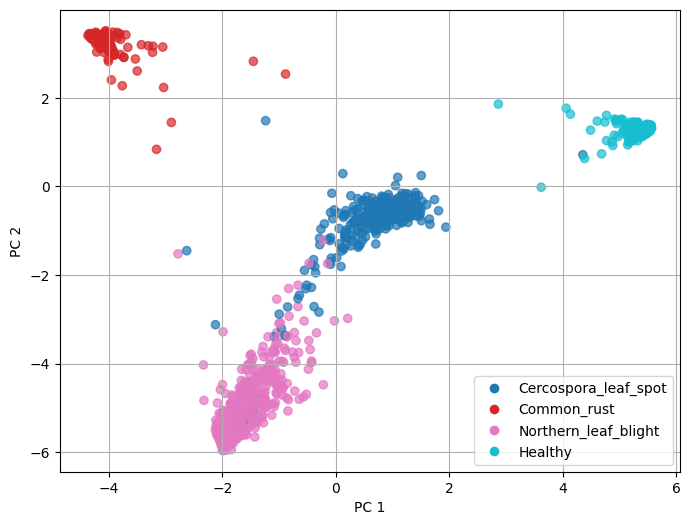}
    \subcaption{Corn 2D PCA}
    \label{fig:corn_2d_pca}
\end{minipage}
\hfill
\begin{minipage}{0.32\linewidth}
    \centering
    \includegraphics[width=\linewidth]{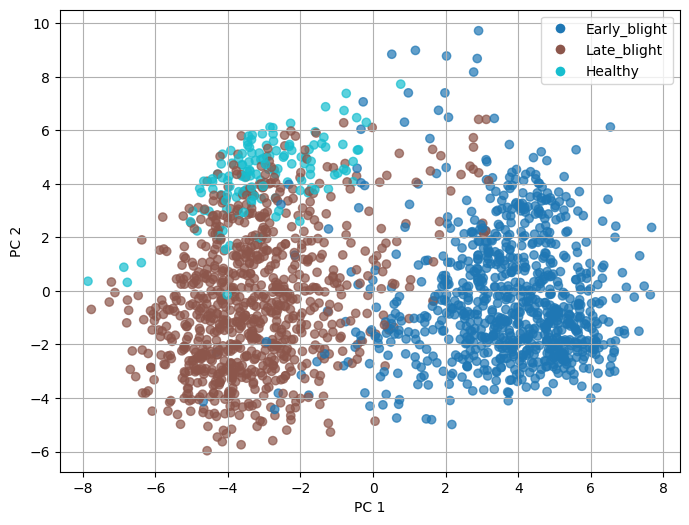}
    \subcaption{Potato 2D PCA}
    \label{fig:potato_2d_pca}
\end{minipage}

\begin{minipage}{0.32\linewidth}
    \centering
    \includegraphics[width=\linewidth]{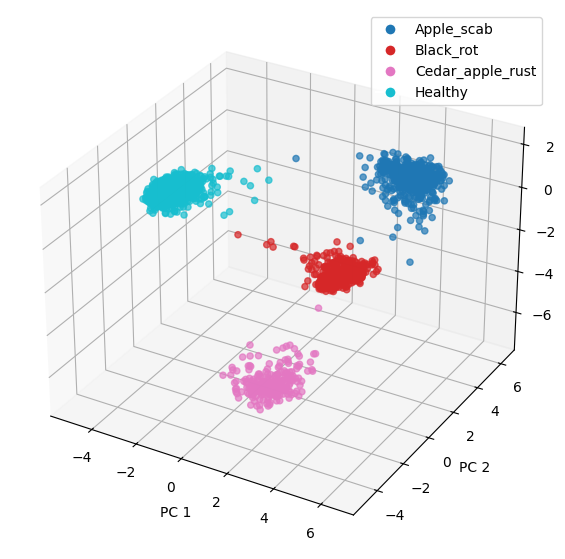}
    \subcaption{Apple 3D PCA}
    \label{fig:apple_3d_pca}
\end{minipage}
\hfill
\begin{minipage}{0.32\linewidth}
    \centering
    \includegraphics[width=\linewidth]{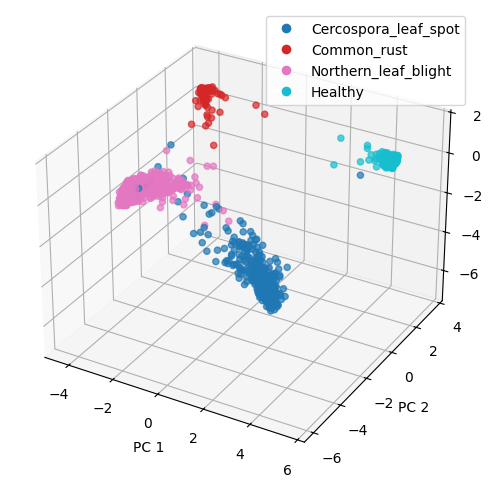}
    \label{fig:corn_3d_pca}
    \subcaption{Corn 3D PCA}
    
\end{minipage}
\hfill
\begin{minipage}{0.32\linewidth}
    \centering
    \includegraphics[width=\linewidth]{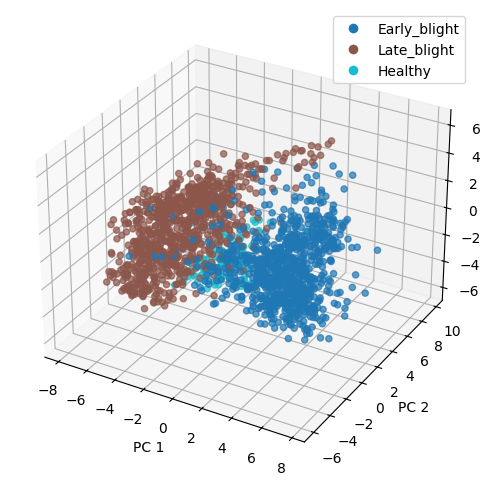}
    \subcaption{Potato 3D PCA}
    \label{fig:potato_3d_pca}
\end{minipage}

\caption{Representative 2D and 3D PCA projections of learned feature distributions for the Apple, Corn, and Potato datasets from PlantVillage}
\label{fig:2d_3d_pca_crops}
\end{figure*}

\textbf{\textit{Principal Component Analysis Visulaizaton:}}
Principal Component Analysis (PCA) is a widely used linear dimensionality reduction technique that projects high-dimensional feature representations into a lower-dimensional subspace while preserving maximum variance in the data \cite{abdi2010principal}. In this study, we extract 32-dimensional feature vectors from the penultimate layer of the proposed Mam-App model and project them onto the top two and three principal components to visualize class-wise feature distributions and shown in Figure \ref{fig:2d_3d_pca_crops}.

As shown in the PCA visualizations, the Apple dataset exhibits clear clustering behavior, where the four disease classes are largely separated into distinct regions in the 2D space, with only a few negligible overlaps that may be separable along higher dimensions. A similar trend is observed in the 3D PCA visualization, indicating strong inter-class separability, which is consistent with the superior classification performance achieved on the Apple dataset.

For the Corn dataset, most classes form well-defined clusters in both 2D and 3D PCA spaces; however, a small degree of overlap is observed, particularly between Corn Northern Leaf Blight and Corn Cercospora Leaf Spot in the 2D projection. This overlap suggests partial feature similarity between these disease patterns, which aligns with the slightly reduced performance compared to the Apple dataset. 

In contrast, the Potato dataset shows noticeable overlap among classes in both 2D and 3D PCA visualizations, indicating weaker class separability in the learned feature space. This increased overlap correlates with the comparatively lower classification accuracy observed for Mam-App and its feature-based Random Forest and XGBoost variants, as reported in Table~\ref{tab:mamba_app_results_percent}. Overall, the PCA analysis demonstrates that the degree of class separability in the learned feature space directly reflects the classification performance across different crop datasets.

\begin{figure*}[!t]
\centering
\begin{minipage}{0.32\linewidth}
    \centering
    \includegraphics[width=\linewidth]{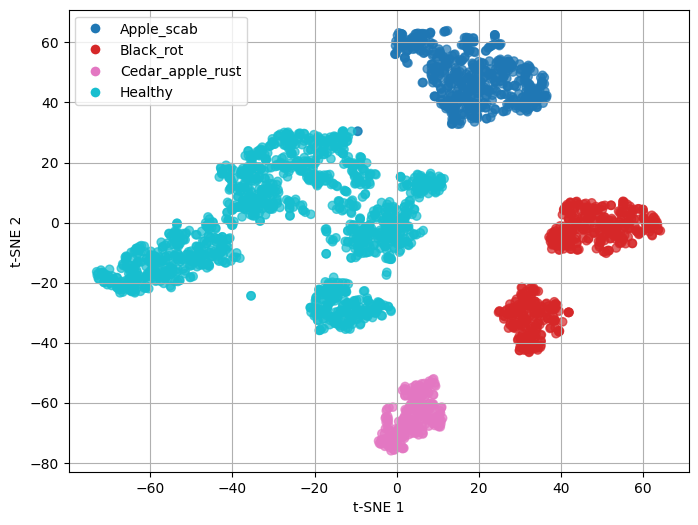}
    \subcaption{Apple 2D t-SNE}
    \label{fig:apple_2d_tsne}
\end{minipage}
\hfill
\begin{minipage}{0.32\linewidth}
    \centering
    \includegraphics[width=\linewidth]{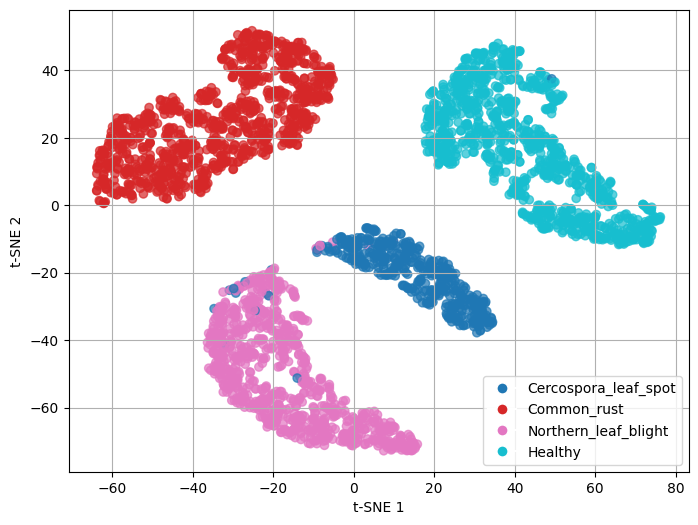}
    \subcaption{Corn 2D t-SNE}
    \label{fig:corn_2d_tsne}
\end{minipage}
\hfill
\begin{minipage}{0.32\linewidth}
    \centering
    \includegraphics[width=\linewidth]{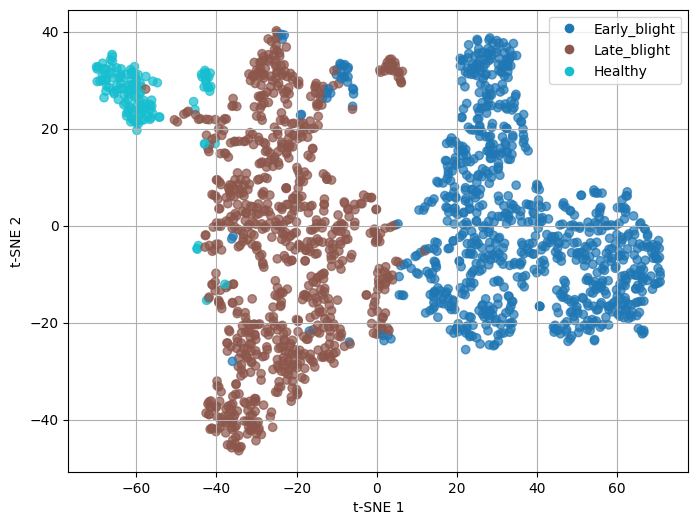}
    \subcaption{Potato 2D t-SNE}
    \label{fig:potato_2d_tsne}
\end{minipage}

\begin{minipage}{0.32\linewidth}
    \centering
    \includegraphics[width=\linewidth]{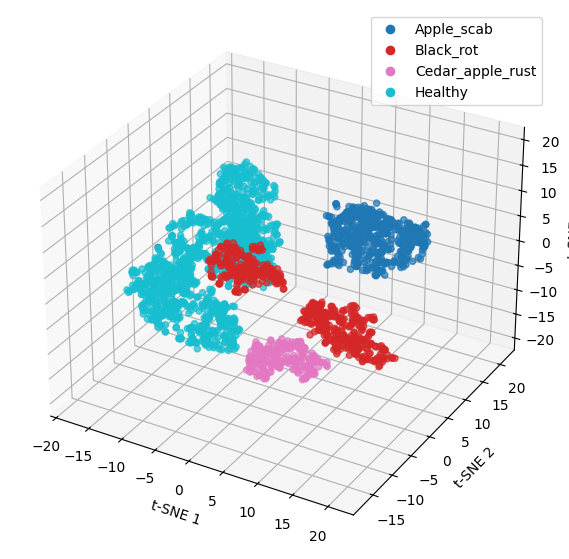}
    \subcaption{Apple 3D t-SNE}
    \label{fig:apple_3d_tsne}
\end{minipage}
\hfill
\begin{minipage}{0.32\linewidth}
    \centering
    \includegraphics[width=\linewidth]{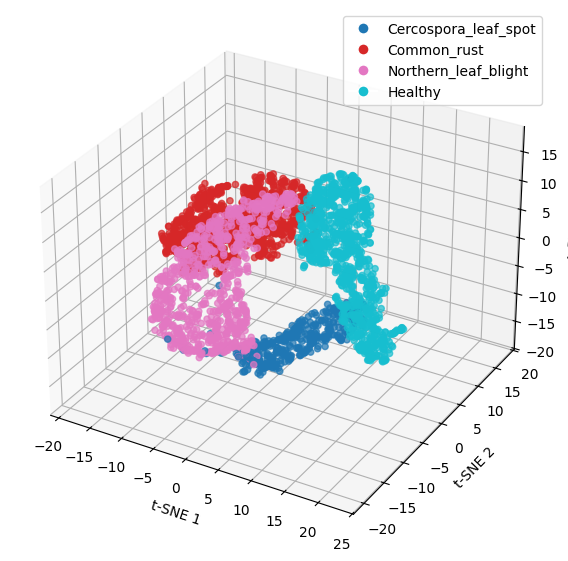}
    \subcaption{Corn 3D t-SNE}
    \label{fig:corn_3d_tsne}
\end{minipage}
\hfill
\begin{minipage}{0.32\linewidth}
    \centering
    \includegraphics[width=\linewidth]{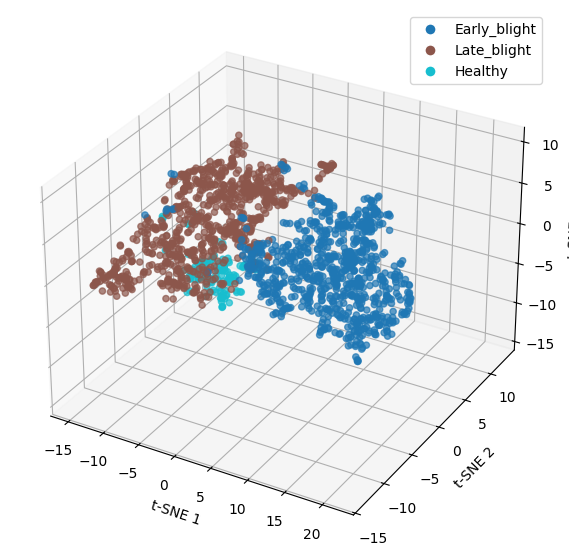}
    \subcaption{Potato 3D t-SNE}
    \label{fig:potato_3d_tsne}
\end{minipage}

\caption{2D and 3D t-SNE visualizations of feature distributions for the Apple, Corn, and Potato datasets from PlantVillage.}
\label{fig:2d_3d_tsne_crops}
\end{figure*}

\textbf{\textit{t-SNE-based feature visualization:}} To further analyze the discriminative capability of the learned features in a non-linear embedding space, we employ t-distributed Stochastic Neighbor Embedding (t-SNE), a popular technique for visualizing high-dimensional data by preserving local neighborhood structures \cite{maaten2008visualizing}. Unlike PCA, t-SNE is particularly effective in revealing complex, non-linear relationships between feature clusters, making it suitable for assessing fine-grained class separability and shown in Figure \ref{fig:2d_3d_tsne_crops}.

In the Apple dataset, both 2D and 3D t-SNE visualizations reveal four compact and well-separated clusters corresponding to the disease classes, demonstrating the strong discriminative power of the features learned by Mam-App. Similarly, the Corn dataset exhibits four clearly distinguishable clusters, with only minor overlaps between visually similar disease categories, further confirming the robustness of the extracted representations. In contrast, the Potato dataset shows comparatively higher overlap between class clusters in both 2D and 3D t-SNE spaces, indicating increased feature ambiguity among disease classes. This observation is consistent with the quantitative results, where the Potato dataset yields slightly lower accuracy across Mam-App, Mam-App + Random Forest, and Mam-App + XGBoost models. Collectively, the t-SNE visualizations provide strong qualitative evidence that Mam-App learns highly discriminative and generalizable feature representations, while also explaining the observed performance variations across different crop datasets.

\section{Discussion}
The experimental results across the Apple, Corn, and Potato PlantVillage disease datasets demonstrate that the proposed Mam-App model achieves highly competitive performance (Tables~\ref{tab:comparison} and~\ref{tab:mamba_app_results_percent}) while maintaining an exceptionally low computational cost. Compared with traditional convolutional networks such as MobileNetV2 and Deep CNNs, as well as recent transformer-based and hybrid architectures, Mam-App consistently delivers higher or comparable accuracy, precision, recall, and F1-score despite using significantly fewer parameters.

A key strength of Mam-App lies in its parameter-efficient architecture, which effectively balances local and global feature learning. The initial convolutional stem blocks are responsible for capturing fine-grained local spatial patterns from leaf images, such as texture variations, vein structures, and disease-specific spots. These local features are then processed by the Mamba-based state-space layers, which act as an efficient alternative to attention mechanisms by modeling long-range dependencies with linear complexity. This design enables Mam-App to capture both local disease characteristics and global contextual information without incurring the heavy computational overhead typically associated with transformer-based models.

Furthermore, the use of global average pooling prior to classification helps retain the most discriminative features while reducing feature dimensionality, contributing to improved generalization and robustness. The final fully connected layer projects a compact $32$-dimensional representation to the target disease classes, further reinforcing the model’s lightweight nature. This architectural choice is reflected in the extremely low parameter count of only 0.051M, making Mam-App well suited for deployment on resource-constrained platforms such as mobile devices, drones, and edge-based agricultural monitoring systems.

The strong generalization capability of Mam-App is further validated through cross-dataset evaluation on the Corn and Potato PlantVillage datasets. Despite differences in disease appearance and inter-class similarity, the model maintains high classification performance, indicating that the learned representations are not dataset-specific but transferable across crop types. The slightly lower performance on the Potato dataset can be attributed to higher visual similarity between disease classes, yet Mam-App still achieves robust accuracy and F1-score compared to larger baseline models.

In addition, PCA and t-SNE reveals clear class-wise clustering, highlighting the discriminative power of the learned feature space. This observation is further supported by downstream classification experiments using Random Forest and XGBoost trained on features extracted from the penultimate layer of Mam-App. The strong performance of these classical classifiers confirms that the Mamba-based representations are informative, well-structured, and suitable for both end-to-end deep learning and hybrid inference pipelines.

Overall, these findings demonstrate that Mam-App effectively addresses the trade-off between performance and efficiency that commonly limits lightweight models. By combining convolutional feature extraction with Mamba-based state-space modeling, Mam-App provides a scalable, robust, and deployable solution for plant disease recognition, making it a practical choice for real-world precision agriculture applications.

\section{Conclusion and Future Work}

This study presents Mam-APP, a parameter-efficient hybrid Mamba-based model for apple leaf disease classification. To demonstrate the robustness and generalization capability of the proposed approach, we further evaluated Mam-APP on the PlantVillage Corn and Potato leaf disease datasets as ablation studies. Despite its extremely compact design with only 0.051M parameters, Mam-APP achieves performance comparable to or better than several state-of-the-art deep learning models. The low parameter count enables fast inference, reduced memory consumption, and low latency, making the model well suited for deployment on resource-constrained devices such as mobile phones, edge devices, and agricultural drones.

Future work will focus on extending Mam-APP toward real-world agricultural applications. First, we plan to explore field-level deployment using images captured under natural conditions, including varying illumination, complex backgrounds, and multiple disease stages, to further validate the model’s robustness. Additionally, incorporating more diverse crop datasets and seasonal variations will improve the generalizability of the proposed framework. Integration with early-warning and decision-support systems can further enhance its practical impact by enabling timely disease detection and intervention.

Second, we aim to develop a mobile application integrated with an LLM-based conversational chatbot, allowing farmers to interactively receive disease diagnosis, treatment recommendations, and guidance on appropriate pesticide or spray usage. The chatbot may also facilitate direct communication with agricultural experts or local extension services, bridging the gap between advanced AI technologies and end users in precision agriculture.

\bibliographystyle{unsrt}  
\bibliography{references}  

\end{document}